\documentclass[10pt,twocolumn,letterpaper]{article}

\usepackage[pagenumbers]{cvpr} 

\definecolor{checkmark}{HTML}{305AFF}
\definecolor{xmark}{HTML}{E62020}

\definecolor{cvprblue}{rgb}{0.21,0.49,0.74}
\usepackage[pagebackref,breaklinks,colorlinks,allcolors=cvprblue]{hyperref}

\def\confName{CVPR}
\def\confYear{2026}
\def\paperID{*****} 

\usepackage{algorithm}

\usepackage[table]{xcolor}
\usepackage{amssymb}
\usepackage{amsmath}
\usepackage{graphicx}
\usepackage[dvipsnames]{xcolor}
\usepackage{algorithm}
\usepackage{algpseudocode}
\usepackage{multirow}
 \usepackage{color}
 \usepackage{amssymb}
\usepackage{amsmath}
\usepackage{graphicx}
\usepackage{colortbl}
\usepackage{color}
\definecolor{mygray}{gray}{.9}
\usepackage{booktabs}
\usepackage[utf8]{inputenc}
\usepackage{url}
\usepackage{booktabs}
\usepackage{amssymb}
\usepackage{bbding}
\usepackage{pifont}
\usepackage{wasysym}
\usepackage{utfsym}
\usepackage{fontawesome}

\title{Partial Ring Scan: Revisiting Scan Order in Vision State Space Models}

\author{Yi-Kuan Hsieh\\
College of Artificial Intelligence \\  National Yang Ming Chiao Tung University\\
\and
Kuan-Chuan Peng\\
Mitsubishi Electric Research Labs, U.S.A\\
\and
Xin li\\
Computer Science Department \\ University at Albany, SUNY, NY, USA. \\
\and
Ming-Ching Chang \\
Computer Science Department\\ University at Albany, SUNY, NY, USA. \\
\and
Yu-Chee Tseng\\
College of Artificial Intelligence \\ National Yang Ming Chiao Tung University\\
\and
Jun-Wei Hsieh\\
College of Artificial Intelligence \\ National Yang Ming Chiao Tung University\\
}

\begin{document}

\maketitle

\begin{abstract}

State Space Models (SSMs) have emerged as efficient alternatives to attention for vision tasks, offering linear-time sequence processing with competitive accuracy. Vision SSMs, however, require serializing 2D images into 1D token sequences along a predefined scan order, a factor often overlooked. We show that scan order critically affects performance by altering spatial adjacency, fracturing object continuity, and amplifying degradation under geometric transformations such as rotation. We present {\bf Partial RIng Scan Mamba (PRISMamba)}, a rotation-robust traversal that partitions an image into concentric rings, performs order-agnostic aggregation within each ring, and propagates context across rings through a set of short radial SSMs. Efficiency is further improved via partial channel filtering, which routes only the most informative channels through the recurrent ring pathway while keeping the rest on a lightweight residual branch. On ImageNet-1K, PRISMamba achieves 84.5\% Top-1 with 3.9G FLOPs and 3,054 img/s on A100, outperforming VMamba in both accuracy and throughput while requiring fewer FLOPs. It also maintains performance under rotation, whereas fixed-path scans drop by 1–2\%. These results highlight scan-order design, together with channel filtering, as a crucial, underexplored factor for accuracy, efficiency, and rotation robustness in Vision SSMs. Code will be released upon acceptance.

\end{abstract}

\begin{figure}[t]
\centerline{
\includegraphics[width=\linewidth]{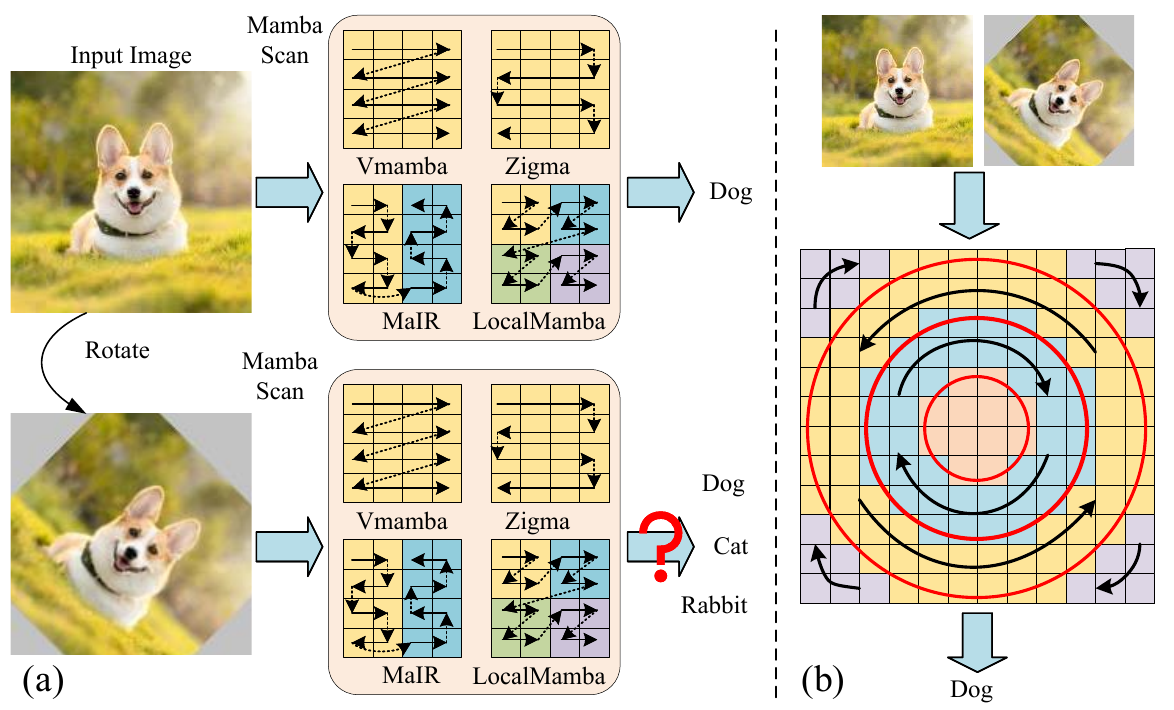}
}
\caption{\textbf{Scanning order affects Vision-Mamba performance.}
\emph{(a) Fixed-path scans} ({\em e.g.}, raster or serpentine in VMamba~\cite{liu2024vmambazhu2024visionmamba}, Zigma~\cite{hu2024zigma}, MaIR~\cite{li2025mair}, LocalMamba~\cite{huang2024localmamba}) preserve sequence–space alignment only under flips. An in-plane rotation (here $60^\circ$) causes padding and global reindexing,  fracturing the path so the recurrent kernel moves along misaligned neighborhoods.
\emph{(b) Our Ring Scan} treats serialization as {\em order-agnostic} aggregation within \emph{concentric rings}, followed by {\em radial} composition from inner to outer, producing a rotation-stable sequence without polar remapping or rotation-specific training.
}
\label{fig:teaser}
\end{figure}

\begin{figure*}[t]
\centerline{
\includegraphics[width=\textwidth]{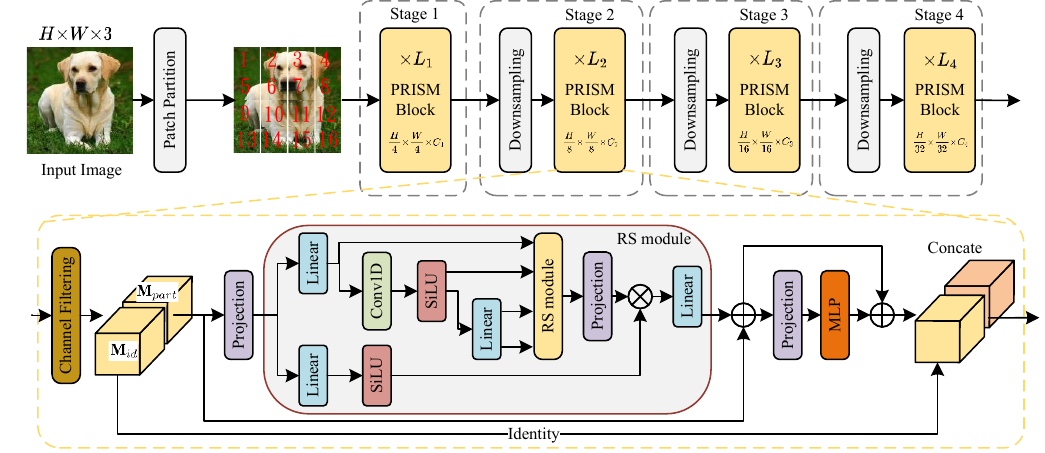}
}
\caption{\textbf{Architecture with Partial RIng Scan Mamba (PRISMamba).}
The image is patchified and processed by a four-stage backbone; stage $i$ stacks $L_i$ \textbf{PRISM} blocks (\emph{Partial RIng Scan Mamba}) with $C_i$ output channels, and stages are separated by downsampling. Each PRISM performs order-agnostic aggregation over a \emph{subset of concentric rings} (partial ring scan), composes information radially with a short sequence operator, and writes features back via a $1{\times}1$ projection before residual fusion. Channel filtering routes only the most informative channels while keeping the rest on a lightweight residual branch for further efficiency improvement.}
\vspace{-0.2in}
\label{arc}
\end{figure*}

\section{Introduction}

Recurrent State Space Models (SSMs) have recently emerged as a competitive alternative to attention-based architectures for long-context vision, offering linear-time sequence processing while maintaining strong accuracy~\cite{gu2021efficiently,gu2024mamba}. Vision adaptations such as Vision Mamba, VMamba~\cite{liu2024vmambazhu2024visionmamba}, and PlainMamba embed selective SSM blocks into hierarchical backbones, which requires {\em serializing} a 2D image into a 1D token sequence according to a chosen {\bf scan order}~\cite{zhu2024visionmambaefficientvisual,liu2024vmambazhu2024visionmamba,yang2024plainmamba}. Although scan order is often treated as a mere implementation detail, for example using row-wise, column-wise, or serpentine traversals, it implicitly defines which spatial neighbors appear adjacent in the sequence. This adjacency governs what local structures the SSM can effectively model with short-range recurrences.

Mounting evidence shows that scan order is far from innocuous. Large-scale studies demonstrate that patch ordering can produce statistically significant performance differences, sometimes exceeding ten Dice points in segmentation tasks~\cite{hardan2025flatten}. In remote sensing, experiments indicate that complex multi-directional scans do not consistently outperform simpler rasterizations, which can suffice depending on the domain~\cite{zhu2024rethinking}. These observations suggest that scan order should be treated as a first-class, cost-free hyperparameter that mediates the alignment between {\em sequence adjacency} (what the SSM processes) and {\em geometric adjacency} (the true image structure).

In this paper, we examine the role of scan order in modern Vision State Space Models (SSMs) and use this analysis to motivate a new traversal and architectural redesign. Conventional SSM traversal patterns often disrupt object continuity: when objects extend in directions misaligned with the scan path, their patches become interleaved with unrelated regions, forcing the SSM to spend capacity repairing local coherence rather than modeling semantics. The problem worsens under geometric transformations such as in-plane rotations, which globally reindex the image and amplify discontinuities. We show that scan order can materially affect performance by altering spatial adjacency, fragmenting object structure, and increasing sensitivity to distortions. 

To address this, we propose {\bf Partial RIng Scan Mamba (PRISMamba)}, a rotation-robust traversal scheme that partitions an image into concentric rings, forms compact ring descriptors, and propagates context radially through a selective SSM. Fig.~\ref{fig:teaser} illustrates the challenge and our solution. The top row shows that common scan paths (raster, bidirectional, diagonal, serpentine) are brittle under rotations, which disrupt token adjacency and break object continuity. In contrast, Fig.~\ref{fig:teaser}(b) shows our ring-based traversal: the image is decomposed into concentric rings, each summarized into a stable descriptor, and these descriptors are processed from inner to outer by a lightweight radial SSM. This preserves spatial locality without relying on angle-specific ordering. When paired with an object detector such as the YOLO family \cite{jocher2024yolov11, tian2025yolov12}, Ring Scan can also be extended to object-aware SSMs, further improving alignment between traversal structure and object geometry.

We evaluate across classification, detection, and segmentation tasks. On ImageNet-1K (224$\times$224), PRISMamba achieves 84.5\% Top-1 with 3.9G FLOPs and 3,054 img/s on A100 GPU, outperforming VMamba~\cite{liu2024vmambazhu2024visionmamba} (82.6\%, 5.6G, 1,686 img/s) by 1.9\% while using 30\% fewer FLOPs and roughly 1.8$\times$ higher throughput. A PRISMamba variant without channel filtering attains 84.1\% at 4.6G and 2,177 img/s, showing that the \emph{Partial} module reduces FLOPs and improves accuracy. On MS~COCO (1$\times$ schedule, 1280$\times$800), PRISMamba attains 48.9 AP$^{\text{box}}$ and 43.2 AP$^{\text{mask}}$ with 235G FLOPs, outperforming VMamba (46.5/42.1 at 262G) and GroupMamba (47.6/42.9 at 279G) while using 10--19\% less FLOPs. Because the token count is unchanged and each ring uses only short recurrences, memory and runtime remain close to standard Vision-SSMs, yet the design delivers clear gains in accuracy and rotation robustness.

Our main contributions are summarized as follows:

\begin{itemize}
\item \textbf{SSM Scan-order Analysis:} We provide the first systematic study of how traversal paths shape spatial adjacency in Vision SSMs. We show that common raster and serpentine scans can fracture object continuity and are highly sensitive to geometric transformations such as rotation.

\item \textbf{Ring Scan for Vision SSMs:} Building on this analysis, we introduce a rotation-stable scan scheme that groups pixels into concentric rings, performs order-agnostic aggregation within each ring, and propagates information radially through short selective SSMs. This preserves the linear-time SSM core while avoiding the fragility of global path-based serialization.

\item \textbf{Partial Channel Filtering (PCF):} We enhance efficiency by forwarding only the most informative channels through the ring-wise recurrent path and routing the remainder through a lightweight residual branch. This improves throughput and reduces FLOPs while modestly improving accuracy.

\item \textbf{Unified architecture and empirical results:} Integrating the above ideas yields PRISMamba, achieving state-of-the-art performance among Vision SSMs. It improves accuracy and throughput on ImageNet-1K and COCO while using fewer FLOPs, and it substantially increases robustness to rotation without any rotation-specific training.

\end{itemize}

\section{Related Work}

\subsection{Convolutional Neural Networks (CNNs)}

Convolutional neural networks (CNNs) have been the cornerstone of visual recognition since AlexNet~\cite{krizhevsky2012imagenet}. Extensive research has continuously advanced their modeling power~\cite{simonyan2014very, szegedy2015going, he2016deep, huang2017densely} and computational efficiency~\cite{howard2017mobilenets, tan2019efficientnet, yang2021focal, radosavovic2020designing} across diverse vision tasks. Advanced operators such as depthwise~\cite{howard2017mobilenets} and deformable convolutions~\cite{dai2017deformable, zhu2019deformable} further improved flexibility and representational capacity.

More recent efforts, inspired by the success of Transformers~\cite{vaswani2017attention}, modern CNNs~\cite{liu2022convnet} incorporate long-range dependencies~\cite{ding2022scaling, rao2022hornet, liu2022more} and dynamic weighting mechanisms~\cite{han2021connection}. These hybrid architectures achieve strong accuracy while preserving the inductive biases and efficiency advantages of convolution.

\subsection{Vision Transformer (ViTs)}

The Vision Transformer (ViT)~\cite{dosovitskiy2020image} first demonstrated that a pure Transformer architecture can achieve strong performance on visual recognition, highlighting the importance of large-scale pre-training. To mitigate ViT’s reliance on massive datasets, DeiT~\cite{touvron2021training} introduced a teacher–student distillation strategy that transfers CNN inductive biases to ViTs. Building on this foundation, numerous works proposed hierarchical and locally aware variants~\cite{liu2021swin, dong2022cswin, wang2021pyramid, NEURIPS2021_9fe77ac7, zhang2023hivit, tian2023integrally, dai2021coatnet, ding2022davit, zhao2022graformer, ali2021xcit} that improve scalability and efficiency.

Another major research direction aims to alleviate the quadratic complexity of self-attention. Linear Attention~\cite{katharopoulos2020transformers} reformulates attention as a linear dot product of kernel feature maps, reducing computational cost from quadratic to linear. GLA~\cite{yang2023gated} introduces a hardware-friendly design that balances memory access and parallelism. RWKV~\cite{peng2023rwkv} integrates linear attention with RNN-style inference, enabling parallelizable training while preserving recurrent efficiency. RetNet~\cite{sun2023retentive} incorporates gating to construct a fully parallelizable alternative to recurrence, while RMT~\cite{Fan2024CVPR} extends this paradigm to vision by adapting temporal decay mechanisms into the spatial domain for representation learning.

Recent ViT research has aimed to improve computational efficiency and better capture spatial dependencies. SHViT~\cite{yun2024shvit} integrates single-head self-attention with convolutional layers to reduce redundancy in early stages, achieving faster inference and higher accuracy on GPUs and mobile devices. GCViT~\cite{Ali2023learning} combines global and local attention to handle multi-scale spatial interactions, yielding strong results in classification and segmentation. Scale-Aware Modulation Transformer (SMT)~\cite{LinICCV2023} employs multi-head mixed convolution and scale-aware aggregation to model the transition from shallow to deep dependencies, achieving notable accuracy improvements. TransNeXt~\cite{Dai2024CVPR} introduces biomimetic foveal attention for efficient visual processing and information fusion with fewer parameters.

\subsection{State Space Models (SSMs)}

While Vision Transformers achieve strong performance, their quadratic attention cost limits scalability to high-resolution inputs. State Space Models (SSMs) have emerged as efficient alternatives, offering linear-time sequence modeling with strong long-range dependency capture~\cite{dao2022flashattention, dao2023flashattention, peng2023rwkv, sun2023retentive, ma2022mega}. HiPPO initialization~\cite{gu2020hippo} enables SSMs to model long sequences effectively, and the S4 framework~\cite{gu2021efficiently} improves efficiency through normalized diagonal parameterization. Building on this, structured variants have been proposed, including complex-diagonal parameterizations~\cite{gupta2022diagonal, gu2022parameterization}, multi-input multi-output extensions~\cite{smith2022simplified}, diagonal-plus-low-rank decompositions~\cite{hasani2022liquid}, and adaptive selection mechanisms~\cite{gu2024mamba}, which have been incorporated into large-scale models~\cite{mehta2022long, ma2022mega, fu2022hungry}.

Although SSMs have excelled in text and speech, their use in vision remains underexplored. Recent work begins to close this gap by adapting Mamba-style SSMs to images, balancing linear-time sequence modeling with 2D inductive biases in scanning, frequency, and architecture. Efficiency-focused designs include \textit{Adventurer}~\cite{wang2025adventurer}, which optimizes Vision-Mamba backbones for faster training without accuracy loss, and \textit{TinyViM}, which uses hybrid Conv–Mamba blocks to capture low-frequency content via a Laplace-domain mixer~\cite{wang2025adventurer, ma2025tinyvim}.

\begin{figure*}[t]
\centerline{
\includegraphics[width=\textwidth]{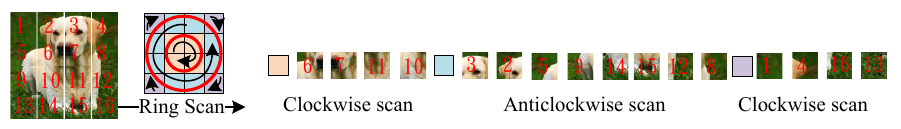}
\vspace{-3mm}
}
\caption{\textbf{Ring Scan.} Pixels are partitioned into concentric rings, which are interactively traversed in a clockwise or counterclockwise sequence. The resulting features are then aggregated in an order-independent fashion, proceeding from inner to outer rings.}
\vspace{-0.2in}
\label{fig:ring:scan}
\end{figure*}

\begin{figure*}[t]
\centerline{
  \includegraphics[width=\textwidth]{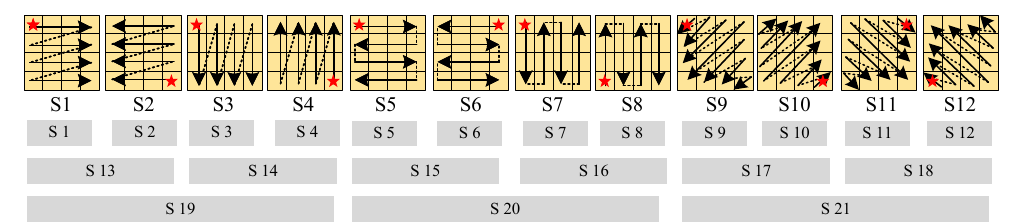}
\vspace{-2mm}
}
\caption{\textbf{Primitive scan orders.} Twelve canonical paths (S1–S12) such as left-to-right raster, serpentine, and diagonal produce distinct 1D sequences from the same image. S1-12 evaluate single scans; S13–18 evaluate pairs of scans; S~19–21 aggregate four scans, enabling a systematic comparison of scan-order effects.}
\label{HVScan}
\end{figure*}

Scan-path design is a key focus. \textit{PlainMamba} uses continuous 2D scanning with direction-aware updates for non-hierarchical recognition~\cite{yang2024plainmamba}; \textit{FractalMamba} employs fractal curves to better preserve multi-scale neighborhoods~\cite{xiao2025boosting}; \textit{ZigMa} applies DiT-style zigzag scanning to diffusion, improving speed and memory at high resolution~\cite{hu2024zigma}. Beyond fixed paths, \textit{DefMamba} learns deformable, content-adaptive scans to prioritize salient structures~\cite{liu2025defmamba}, while \textit{VSSD} leverages bidirectional visual context via non-causal SSMs~\cite{shi2025vssd}. For restoration, \textit{MaIR} enforces locality and continuity using nested S-shaped routes with lightweight fusion, achieving state-of-the-art results on structure-sensitive benchmarks~\cite{li2025mair}.

Other refinements include \textit{Mamba-Reg}, which inserts evenly spaced register tokens to stabilize scaling and suppress high-norm background artifacts~\cite{wang2025mamba}. Task-specific extensions demonstrate Mamba’s versatility: \textit{Selective Visual Prompting} streamlines adaptation of Vision-Mamba backbones, and \textit{Mamba as a Bridge} fuses foundation vision and vision-language priors for domain-generalized semantic segmentation, achieving competitive mIoU~\cite{yao2025selective, zhang2025mamba}.

\subsection{Scan-Based Vision SSMs}


Recent Vision Mamba architectures adapt selective state-space blocks to visual data by \emph{scanning} 2D feature grids to define a 1D token processing order, along with the SSM propagates states. VMamba~\cite{liu2024vmambazhu2024visionmamba} integrates a 2D Selective Scan (SS2D) into a hierarchical backbone, achieving strong accuracy–efficiency trade-offs across both classification and dense prediction tasks~\cite{liu2024vmambazhu2024visionmamba}. For medical and remote-sensing segmentation, VM-UNet replaces or augments attention with Visual State-Space blocks in a U-shaped design, improving long-range dependency modeling at low computational cost~\cite{ruan2024vm}.

Beyond single-path rasterization, 2DMamba explicitly incorporates 2D spatial structure through selective operators, improving representation quality on both natural and whole-slide imagery~\cite{zhang20252dmamba}. Recent surveys~\cite{zhang2024survey, liu2025vision} systematically categorize these Vision Mamba variants, identifying emerging design patterns such as multi-directional or axial scans, tiled or windowed scanning, and lightweight positional priors. Hardware-aware adaptations further streamline selective scanning for mobile or edge deployment~\cite{pei2025efficientvmamba}.

Despite this progress, most pipelines still rely on \emph{flip-only} augmentation to preserve scan continuity. When inputs undergo arbitrary rotations, fixed-path traversals suffer from boundary padding and reindexing effects that disrupt token adjacency, which highlights the need for traversal schemes that remain geometrically consistent under rotation.

\section{Method}
\label{sec:method}

\subsection{From SSMs to Vision SSMs}
\label{subsec:ssm-to-vssm}

Let $\{x_k\}_{k=1}^{T}$ be a token sequence with token width $m\in\mathbb{N}$ and length $T\in\mathbb{N}$. A small projector $\Pi(x_k)$ produces stepwise parameters $A_k,B_k\in\mathbb{R}^{d}$ (diagonal) and a  mixing matrix $C_k\in\mathbb{R}^{d\times d}$. With $\odot$ denoting the Hadamard product, a selective SSM maintains a latent state $h_k\in\mathbb{R}^{d}$ and output $y_k\in\mathbb{R}^{d}$ as: 
\begin{equation}
h_k = A_k \odot h_{k-1} + B_k \odot x_k,\text{ }
y_k \,=\, C_k\,h_k,
\label{eq:ssm-core}
\end{equation}
and $h_0=0$. Vision SSMs (VSSMs) integrate this sequence operator into a hierarchical backbones by first \emph{serializing} a 2D feature map $X\in\mathbb{R}^{H\times W\times C}$ into a 1D sequence according to a chosen \emph{scan order}. After the SSM processes the sequence, the results are written back to the grid through a $1{\times}1$ projection. The scan order therefore becomes a key design choice, since it determines which spatial neighbors are treated as adjacent in sequence space.

\subsection{Scan Orders in Vision SSMs}
\label{subsec:scan-orders}

A scan order assigns each grid location $(u,v)$ a unique visit index $k\in\{1,\ldots,N\}$, where $N=HW$, thereby linearizing the grid. 
Fig.~\ref{HVScan} illustrates twelve primitive orders ({\em e.g.}, raster, serpentine, diagonal) and their composites that we use to study scan effects. These continuous paths generally perform well under horizontal or vertical flips, as sequence adjacency aligns with spatial neighbors. However, in-plane rotations permute neighborhood relations and introduce padded corners, breaking the correspondence between the 1D dynamics of Eq.~\eqref{eq:ssm-core} and the underlying 2D structure. This motivates a traversal that maintains stable groupings under rotation while keeping sequences short to preserve linear-time efficiency.

\subsection{Ring-by-Ring Alternating Scan}
\label{subsec:ring-scan}

We group pixels by their Euclidean distance to the image center $(c_x,c_y)$ and visit the grid \emph{ring by ring} as in Fig.~\ref{fig:ring:scan}. For a pixel $(u,v)$ with $u \in \{0,\ldots,W{-}1\}$ and $v \in \{0,\ldots,H{-}1\}$, define its radius $r(u,v)=\big\|(u{-}c_x,\ v{-}c_y)\big\|_2.$ Given a ring width $\Delta r>0$, assign the integer ring index:
\begin{equation}
\hat r(u,v)=\Big\lfloor \frac{r(u,v)}{\Delta r}\Big\rfloor,\;\;
\mathcal{P}_r=\{(u,v):\hat r(u,v)=r\}.
\end{equation}
In real implementations, if an object detector is performed first, $(c_x,c_y)$ can be an object's center to make our method an object-aware descriptor.

\paragraph{Alternating loop order:}
Within each ring $r$, we construct a simple closed loop $\sigma_r:\{1,\ldots,L_r\}\!\to\!\mathcal{P}_r$ (any fixed contour rule works) and \emph{alternate the traversal direction across rings}: clockwise for odd $r$ and counterclockwise for even $r$. 
Let $P\in\mathbb{R}^{m\times C}$ be a $1{\times}1$ projection. The ring-$r$ token sequence is:
\begin{equation}
\label{eq:ring-tokens}
x_{r,k}=P\,X_{\sigma_r(k)}\in\mathbb{R}^{m},\qquad k=1,\ldots,L_r,
\end{equation}
processed in the chosen direction. A lightweight selective SSM runs along the loop:
\begin{equation}
\label{eq:ssm-angular}
\begin{aligned}
h_{r,k}&=A_{r,k}\odot h_{r,k-1}+B_{r,k}\odot x_{r,k},\\
y_{r,k}&=C_{r,k}\,h_{r,k},\qquad h_{r,0}=0,
\end{aligned}
\end{equation}
and the ring descriptor is the average of all per-step outputs:
\begin{equation}
\label{eq:ring-desc}
z_r=\frac{1}{L_r}\sum_{k=1}^{L_r} y_{r,k}\ \in\ \mathbb{R}^{d}.
\end{equation}

\paragraph{Rotation behavior:}
In-plane rotations leave each pixel’s ring index $\hat r(u,v)$ unchanged. As a result, a ring’s loop experiences only a cyclic shift, and alternating traversal directions across rings reduce start-index bias. This design achieves rotation robustness without relying on polar remapping or rotation-specific augmentation.


\subsection{Inner to Outer Radial Ring Scan}
\label{subsec:radial}

The per-ring descriptors $\{z_r\}_{r=0}^{R^\star}$ form a short radial sequence from the innermost to the outermost ring. Context is propagated inward-to-outward using a second selective SSM:
\begin{equation}
\label{eq:ssm-radial}
\begin{aligned}
h^{\mathrm{rad}}_{r}&=\tilde{A}_{r}\odot h^{\mathrm{rad}}_{r-1}+\tilde{B}_{r}\odot z_r,\\
y^{\mathrm{rad}}_{r}&=\tilde{C}_{r}\,h^{\mathrm{rad}}_{r},\qquad h^{\mathrm{rad}}_{-1}=0.
\end{aligned}
\end{equation}
This two-level design, consisting of a per-ring loop in Eq.~\eqref{eq:ssm-angular} and a short radial chain in Eq.~\eqref{eq:ssm-radial}, maintains linear-time complexity: $O(L_r)$ per ring and $O(R^\star{+}1)$ across rings. With practical choices of $\Delta r$, the number of rings $R^\star\!\ll\!\sqrt{HW}$, making the additional recurrence cost  negligible compared to $O(HW)$ for scanning the entire grid.

\subsection{Partial Channel Filtering}
\label{subsec:chfilter}
Before ring processing, we introduce a lightweight \emph{hard routing} over channels to curb the per-channel cost of Ring Scan Module. Feature channels often exhibit unequal salience—many are weak or redundant yet would still incur full recurrent updates if all channels were forwarded. Unlike channel-attention methods such as SE~\cite{hu2018squeeze} and CBAM~\cite{woo2018cbam}, which learn \emph{soft} importance weights but continue to process \emph{all} channels through downstream blocks, our filter quickly separates informative channels from residual ones and only the informative subset enters the ring pathway. This maintains accuracy while substantially reducing FLOPs and activation memory.

Concretely, given a feature map $\mathbf{X}\!\in\!\mathbb{R}^{H\times W\times D_c}$ with channels $\{C_i\}_{i=1}^{D_c}$, we compute per-channel salience via global average pooling (GAP),
$\mu_i=\mathrm{GAP}(C_i),\quad i=1,\dots,D_c,$ and set a mean threshold:
\begin{equation}
\label{eq:mu-thresh}
\mu=\frac{1}{D_c}\sum_{i=1}^{D_c}|\mu_i|.
\end{equation}
A channel is retained if $\mu_i\ge\mu$ and otherwise routed to a residual bypass. Denoting the partition operator by $\Phi$, we obtain:
\begin{equation}
\label{ChannelFilter}
(\mathbf{M}_{\text{part}},\,\mathbf{M}_{\text{id}})=\Phi(\mathbf{X},\mu),
\end{equation}
where $\mathbf{M}_{\text{part}}=\{\,C_i\mid \mu_i\ge\mu\,\}, \mathbf{M}_{\text{id}}=\{\,C_i\mid \mu_i<\mu\,\}$. The informative branch $\mathbf{M}_{\text{part}}$ is fed to the ring-by-ring modules, while $\mathbf{M}_{\text{id}}$ follows an identity/residual path and is fused later. Choosing the mean yields $O(D_c)$ complexity and avoids sorting; a median variant is possible but costs $O(D_c\log D_c)$. In practice, this fast selection preserves accuracy and achieves clear compute savings compared to soft reweighting approaches~\cite{hu2018squeeze,woo2018cbam}, which still bear the full per-channel processing load downstream.

\subsection{Write-Back and Fusion}
\label{subsec:fusion}
Let $\Psi\in\mathbb{R}^{C\times d}$ be a $1{\times}1$ projection. 
For pixel $(u,v)$ on ring $\hat r(u,v)$, we broadcast the ring-level output:
\begin{equation}
\label{eq:writeback}
Y_{u,v}=\Psi\,y^{\mathrm{rad}}_{\hat r(u,v)}\in\mathbb{R}^{C}.
\end{equation}
To avoid the vanishing gradient issues, we fuse with the backbone stream using a lightweight residual projection:
\begin{equation}
\label{eq:fuse}
X_{\text{out}}=X_{\text{in}}+\mathrm{Conv}_{1\times1}(Y),
\end{equation}
which preserves spatial size $H\times W$ and channel size $C$.
An optional pointwise FFN may follow as in standard Vision-Mamba blocks.

\section{Experiment Results}

{\bf Implementation details:}
Our models were trained from scratch using the AdamW optimizer~\cite{loshchilov2017decoupled} over 300 epochs with batch size of 128. The training regime included a linear warm-up during the first five epochs, a momentum of 0.9, and a cosine learning rate schedule~\cite{loshchilov2016sgdr} with an initial learning rate of ${1\times 10^{-4}  }$. For fair comparisons, we adopted the same data augmentation techniques as prior work~\cite{touvron2021training}, including mixup~\cite{zhang2017mixup}, random erasing~\cite{zhong2020random}, and auto-augmentation~\cite{cubuk2019autoaugment}. Throughput was measured on an Nvidia A100 GPU.

We compare against fifteen Vision-Mamba families listed in Table~\ref{tab:class}: Vim~\cite{zhu2024visionmambaefficientvisual}, VMamba~\cite{liu2024vmambazhu2024visionmamba}, SiMBA~\cite{patro2024simba}, Zigma~\cite{hu2024zigma}, QuadMamba~\cite{xie2024quadmamba}, LocalMamba~\cite{huang2024localmamba}, FractalMamba~\cite{xiao2025boosting}, Adventurer~\cite{wang2025adventurer}, SparX-Mamba~\cite{lou2025sparx}, EfficientVMamba~\cite{pei2025efficientvmamba}, PlainMamba~\cite{yang2024plainmamba}, GroupMamba~\cite{shaker2025groupmamba}, VSSD~\cite{shi2025vssd}, and DefMamba~\cite{liu2025defmamba}, MaIR~\cite{li2025mair}.

\subsection{ImageNet-1K Results and Efficiency Analysis}
\label{subsec:imgnet-analysis}

Table~\ref{tab:class} compares our variants against recent Vision-Mamba families at $224^2$ and ablates the effect of channel filtering by contrasting \emph{PRISMamba (w/o PCF)} (PRISMamba without partial channel filtering) with \emph{PRISMamba}. Relative to strong baselines of similar computation ({\em e.g.}, VMamba: 5.6G FLOPs, 82.6\% Top-1; SparX-Mamba: 5.2G FLOPs, 83.5\%), our designs shift the accuracy–efficiency frontier. Notably, PRISMamba (w/o PCF) achieves 84.1\% Top-1 at 4.6G FLOPs and 2177 img/s, surpassing VMamba by +1.5 points while using roughly \textbf{18\%} fewer FLOPs and delivering higher throughput. Additional ablation studies on other scanning orders in Fig.~\ref{HVScan} will be provided in the Supplementary Material.

Introducing \textbf{partial channel filtering} yields further gains. PRISMamba preserves only the importance informative channels for ring-based sequence processing and routes the remainder through a lightweight residual branch. This targeted allocation cuts computation from 4.6G to \textbf{3.9G} FLOPs, boosts throughput from 2177 to \textbf{3054} img/s (+40\%), and also improves Top-1 from 84.1\% to \textbf{84.5\%} (+0.4\%). The improvement in accuracy alongside reduced FLOPs indicates that the channel filter does more than prune cost: by steering the recurrent ring pathway toward high-signal channels, it produces cleaner ring descriptors and more effective radial composition. Compared to prior Mamba variants with similar or larger budgets (e.g., GroupMamba: 83.9\% at 7.0G FLOPs; QuadMamba: 81.4\% at 5.5G FLOPs), PRISMamba attains higher accuracy with substantially lower compute and markedly higher realized throughput on A100.

In summary, the ablation isolates the contribution of channel filtering: selecting salient channels for the ring-by-ring traversal simultaneously reduces FLOPs, increases hardware throughput, and improves recognition accuracy. This supports our design choice to treat scan order and channel allocation as coupled levers for efficiency–accuracy optimization in Vision SSMs.

\begin{table}[t]
\caption{Performance comparison on ImageNet-1K all images are of size $224 \times 224$. Throughput(TP) values are measured with on Nvidia A100 GPU.
\vspace{-2mm}
}
\label{tab:class}
\centerline{
\setlength{\tabcolsep}{1mm}
\resizebox{\linewidth}{!}{
\begin{tabular}{lcccc}
\toprule
Model & Params & GFlops & TP(img/s)   & Top-1 Acc\%\\
\midrule
Vim~\cite{zhu2024visionmambaefficientvisual} & 26 M & 5.3 G & 811 & 80.5\\
VMamba~\cite{liu2024vmambazhu2024visionmamba} & 30 M & 5.6 G & 1686 & 82.6\\
SiMBA~\cite{patro2024simba} & 27 M & 5.0 G & - & 84.0\\
Zigma~\cite{hu2024zigma} & 31 M &  5.1 G & - & 82.4 \\
QuadMamba~\cite{xie2024quadmamba} & 31 M & 5.5 G & 1252 & 81.4\\
LocalMamba~\cite{huang2024localmamba} & 26 M & 5.7 G & - & 82.7 \\
FractalMamba~\cite{xiao2025boosting} & 31 M & 4.8 G & - & 83.0 \\
Adventurer~\cite{wang2025adventurer} & 12 M & 4.2 G & 2757 & 78.2 \\
SparX-Mamba~\cite{lou2025sparx} & 27 M & 5.2 G & 1370 & 83.5 \\
EfficientVMamba~\cite{pei2025efficientvmamba} & 33 M & 4.0 G & - & 81.8 \\
PlainMamba~\cite{yang2024plainmamba} & 25 M & 8.1 G & - & 81.6 \\
GroupMamba~\cite{shaker2025groupmamba} & 34 M & 7.0 G & 803 & 83.9 \\
VSSD~\cite{shi2025vssd} & 24 M & 4.5 G & - & 83.7 \\
DefMamba~\cite{liu2025defmamba} & 26 M & 4.8 G & - & 83.5 \\  
MaIR~\cite{li2025mair} & 26 M & 5.4 G & - & 83.1 \\  \hline
PRISMamba (w/o PCF) & 27 M & 4.6 G & 2177 & 84.1\\
PRISMamba & 22 M & \textcolor{red}{\textbf{3.9 G}} & \textcolor{red}{\textbf{3054}} & \textcolor{red}{\textbf{84.5}}\\
\bottomrule
\end{tabular}%
}
\vspace{-2mm}
}
\end{table}

\subsection{Rotation Robustness of Ring vs. Fixed-Path Scans}
\label{subsec:rotation-analysis}

Table~\ref{tab:rotation} evaluates Top-1 accuracy under controlled in-plane rotations ($0^\circ/30^\circ/60^\circ$). Fixed-path Vision-Mamba baselines uniformly degrade by $\approx$1 to 2 points once the image is rotated, {\em e.g.}, VMamba drops from 82.6 to 80.6 at $60^\circ$, GroupMamba from 83.9 to 82.0, and PlainMamba from 81.6 to 79.8. In contrast, our ring-based scans are essentially flat across angles. PRISMamba (w/o PCF) attains 84.1/83.9/83.9 at $0^\circ/30^\circ/60^\circ$, indicating near-perfect rotation stability. Adding \textbf{partial channel filtering} further improves both the non-rotated baseline and the rotated cases: PRISMamba reaches \textbf{84.5} at $0^\circ$ and remains at \textbf{84.3}/\textbf{84.4} under $30^\circ/60^\circ$, surpassing all prior variants at every angle.

Two key observations emerge. (1) The ring-by-ring alternating traversal (odd rings clockwise, even rings counterclockwise) preserves ring membership under rotation and converts global orientation changes into cyclic shifts along each closed ring, which do not harm the loop-aggregated descriptor. This explains the stability of PRISMamba (w/o PCF) relative to fixed-path scans. (2) Partial channel filtering enhances ring descriptors by emphasizing high-signal channels for the recurrent ring path, improving overall accuracy without compromising rotation invariance.
Together, these results support our central claim: \emph{scan-order design}, combined with targeted channel allocation, resolves the sequence–geometry mismatch in path-based Vision SSMs, providing rotation robustness without specialized training.

\begin{table}[t]
\setlength{\tabcolsep}{8pt}
\caption{\textbf{Rotation stress test (Top-1, \%).} We compare fixed-path Vision Mamba variants against our Ring Scan under no rotation and rotations of $30^\circ$/$60^\circ$ rendered on the same canvas.
\vspace{-2mm}
}
\label{tab:rotation}
\centerline{
\setlength{\tabcolsep}{1mm}
\resizebox{\linewidth}{!}{
\begin{tabular}{lccc}
\toprule
Model & No Rot. & $30^\circ$ Rot. & $60^\circ$ Rot. \\
\midrule
Vim~\cite{zhu2024visionmambaefficientvisual}   & 80.5 & 78.6 \;(\,-1.9\,) & 78.6 \;(\,-1.9\,) \\
VMamba~\cite{liu2024vmambazhu2024visionmamba} & 82.6 & 80.7 \;(\,-1.9\,) & 80.6 \;(\,-2.0\,) \\
SiMBA~\cite{patro2024simba} & 84.0 &  83.1\;(\,-0.9\,) &  82.9\;(\,-1.1\,) \\
Zigma~\cite{hu2024zigma} & 82.4 & 81.1\;(\,-1.3\,) & 80.9\;(\,-1.5\,) \\
QuadMamba~\cite{xie2024quadmamba} & 81.4 &  79.6\;(\,-1.8\,) &  79.9\;(\,-1.5\,) \\
LocalMamba~\cite{huang2024localmamba} & 82.7 & 80.9 \;(\,-1.8\,) & 80.9 \;(\,-1.8\,) \\
FractalMamba~\cite{xiao2025boosting} & 83.0 &  81.9\;(\,-1.1\,) &  81.7\;(\,-1.3\,) \\
Adventurer~\cite{wang2025adventurer}  & 78.2 &  76.5\;(\,-1.7\,) &  76.8\;(\,-1.4\,) \\
SparX-Mamba~\cite{lou2025sparx}  & 83.5 &  82.1\;(\,-1.4\,) &  82.3\;(\,-1.2\,) \\
EfficientVMamba~\cite{pei2025efficientvmamba}  & 81.8 &  79.7\;(\,-2.1\,) &  79.9\;(\,-1.9\,) \\
PlainMamba~\cite{yang2024plainmamba}  & 81.6 &  79.6\;(\,-2.0\,) &  79.8\;(\,-1.8\,) \\
GroupMamba~\cite{shaker2025groupmamba}  & 83.9 &  82.1\;(\,-1.8\,) &  82.0\;(\,-1.9\,) \\
VSSD~\cite{shi2025vssd}  & 83.7 &  82.5\;(\,-1.2\,) &  82.6\;(\,-1.1\,) \\
DefMamba~\cite{liu2025defmamba}  & 83.5 &  81.8\;(\,-1.7\,) &  81.6\;(\,-1.9\,) \\ 
MaIR~\cite{li2025mair}  & 83.1 &  81.5\;(\,-1.6\,) &  81.4\;(\,-1.7\,) \\\hline
PRISMamba (w/o PCF) & 84.1 & 83.9 & 83.9 \\
PRISMamba & \textcolor{red}{\textbf{84.5}} & \textcolor{red}{\textbf{84.3}} & \textcolor{red}{\textbf{84.4}} \\
\bottomrule
\end{tabular}%
}
\vspace{-2mm}
}
\end{table}

\subsection{COCO Detection and Instance Segmentation: Balancing Accuracy Efficiency}
\label{subsec:coco-analysis}

Table~\ref{table:Downstream Tasks} reports Mask R-CNN results on MS~COCO (mini-val, 1$\times$ schedule, $1280{\times}800$) and shows that PRISMamba advances both accuracy and efficiency relative to strong Vision-Mamba baselines. With only \textbf{235G} FLOPs, PRISMamba achieves \textbf{48.9} AP$^{\text{box}}$ and \textbf{43.2} AP$^{\text{mask}}$, outperforming VMamba (46.5/42.1 at 262G) by $+2.4$ box AP and $+1.1$ mask AP while using $\sim$10\% fewer FLOPs. Against GroupMamba (47.6/42.9 at 279G) and DefMamba (47.5/42.8 at 268G), PRISMamba delivers consistent gains $+1.3$ AP$^{\text{box}}$ and $+0.3\!\sim\!+0.4$ AP$^{\text{mask}}$ with lower costs.

Gains hold under stricter localization metrics (AP$^{\text{box}}_{75}$: \textbf{52.6} vs.\ 52.1/51.7), showing that the ring-by-ring alternating scan enhances precise box alignment rather than just recall (AP$_{50}$ also improves). Even against FLOP-heavier variants like PlainMamba-Adapter (542G), PRISMamba achieves higher accuracy with less than half the computational cost. These results show that careful scan-order design can deliver tangible improvements on downstream tasks without increasing model size or training schedule.

\begin{table}[t]
\centering
\setlength{\tabcolsep}{2pt}
\caption{Mask R-CNN object detection and instance segmentation on MS COCO mini-val using 1× schedule. FLOPs are computed using input size $1280 \times 800$.
\vspace{-2mm}
}
\label{table:Downstream Tasks}
\resizebox{\linewidth}{!}{
\begin{tabular}{c|c|ccc|ccc}
\hline
 & FLOPs  & \multicolumn{3}{c|}{Object Det.}             & \multicolumn{3}{c}{Instance Seg.}           \\ \cline{3-8} 
\multirow{-2}{*}{Model}      &  (G)           & AP$^{box}$    & AP$^{box}_{50}$ & AP$^{box}_{75}$ & AP$^{mask}$   & AP$^{mask}_{50}$ & AP$^{mask}_{75}$ \\ \hline
Vim                           & -             & 45.7          & 63.9            & 49.6            & 39.2          & 60.9             & 41.7             \\
VMamba                        & 262           & 46.5          & 68.5            & 50.7            & 42.1          & 65.5             & 45.3             \\
QuadMamba                     & 301           & 46.7          & 69.0            & 51.3            & 42.4          & 65.9             & 45.6             \\
LocalMamba                   & 291           & 46.7          & 68.7            & 50.8            & 42.2          & 65.7             & 45.5             \\
FractalMamba                  & 266           & 46.8          & 68.7            & 50.8            & 42.4          & 65.9             & 45.8             \\
Adventurer                    & -             & 46.5          & 65.2            & 50.4            & 40.3          & 62.2             & 43.5             \\
EfficientVMamba               & -             & 41.6          & 63.2            & 45.3            & 38.6          & 60.5             & 41.5             \\
PlainMamba-Adapter           & 542           & 46.0          & 66.9            & 50.1            & 40.6          & 63.8             & 43.6             \\
GroupMamba                    & 279           & 47.6          & 69.8            & 52.1            & 42.9          & 66.5             & 46.3             \\
VSSD                          & 265           & 46.9          & 69.4            & 51.4            & 42.6          & 66.4             & 45.9             \\
DefMamba                      & 268           & 47.5          & 69.6            & 51.7            & 42.8          & 66.3             & 46.2 \\ \hline
PRISMamba & \textcolor{red}{\textbf{235}} & \textcolor{red}{\textbf{48.9}} & \textcolor{red}{\textbf{70.7}}   & \textcolor{red}{\textbf{52.6}}   & \textcolor{red}{\textbf{43.2}} & \textcolor{red}{\textbf{67.4}}    & \textcolor{red}{\textbf{46.8}}    \\ \hline
\end{tabular}
}
\end{table}

\begin{table}[t] 
\caption{Ablation of Partial Channel Filtering across Vision-Mamba families. Throughput (TP) values are measured with on Nvidia A100 GPU.
\vspace{-2mm}
} 
\label{abl:channel filter} 
\centerline{
\setlength{\tabcolsep}{1mm}
\resizebox{\linewidth}{!}{ 
\begin{tabular}{lcccc} 
\toprule Model & Params. & GFlops & TP (img/s) & Top-1 Acc\%\\ 
\midrule Vim & 26 M & 5.3 G & 811 & 80.5\\
+PCF & 23 M & 4.9 G (-0.4)& 1183 & 80.8 (+0.3\%)\\ \hline 
VMamba & 30 M & 5.6 G & 1686 & 82.6\\ 
+PCF & 26 M & 5.1 (-0.5) G & 2314 & 82.9 (+0.3\%)\\ \hline 
QuadMamba & 31 M & 5.5 G & 1252 & 81.4\\ 
+PCF & 25 M & 5.0 G (-0.5) & 1871 & 81.7 (+0.2\%)\\ \hline 
Adventurer & 12 M & 4.2 G & 2757 & 78.2 \\ 
+PCF & 10 M & 4.0 G (-0.2) & 3819 & 78.4 (+0.2\%)\\ \hline 
SparX-Mamba & 27 M & 5.2 G & 1370 & 83.5 \\ 
+PCF & 23 M & 4.8 G (-0.4) & 1959 & 83.9 (+0.4\%)\\ \hline 
GroupMamba & 34 M & 7.0 G & 803 & 83.9 \\ 
+PCF & 30 M & 6.4 G (-0.6) & 1075 & 84.2 (+0.3\%)\\ \hline 
PRISMamba (w/o PCF) & 27 M & 4.6 G & 2177 & 84.1\\ 
PRISMamba & 22 M & 3.9 G (-0.7) & 3054 & 84.5 (+0.4\%)\\ \bottomrule 
\end{tabular}%
} 
\vspace{-2mm}
}
\end{table}


\subsection{Generalization of Partial Channel Filtering}
\label{subsec:ablation-pcf}
Table~\ref{abl:channel filter} examines whether mean-based \emph{Partial Channel Filtering} (PCF) transfers beyond our architecture by inserting it into diverse Vision-Mamba backbones. Across Vim, VMamba, QuadMamba, Adventurer, SparX-Mamba, and GroupMamba, PCF consistently lowers computation and raises accuracy: parameters drop by 2--6M and FLOPs by 0.2--0.6G (roughly 4--12\%), while throughput increases by about 35--50\% and Top-1 improves by +0.2--0.4 pp. These gains are remarkably uniform across capacities and design variants, suggesting that (i) many channels contribute marginally to ring-style recurrent updates, and (ii) forwarding only high-activation channels through the recurrent path produces cleaner descriptors without starving the model of global context (residual branch). 

Our models show the same pattern. PRISMamba (w/o PCF) (84.1\%, 4.6G, 2177 img/s); re-enabling PCF gives PRISMamba (84.5\%, 3.9G, 3054 img/s), i.e., \(-15\%\) FLOPs, \(\sim\!+40\%\) throughput, and \(+0.4\) pp Top-1. Notably, accuracy increases despite reduced compute, indicating that PCF does more than prune cost: it improves the signal-to-noise ratio of ring descriptors and stabilizes the subsequent radial composition. Together, these results support PCF as a simple, general plug-in for Vision-SSMs that advances the accuracy--efficiency frontier without architectural surgery or re-training tricks.

\subsection{Comparing soft channel attention (SE, CBAM) with hard Partial Channel Filtering (PCF)}
Table~\ref{abl:channel filter VS SE} isolates the effect of channel selection strategy on the same backbone.
Relative to PRISMamba (w/o PCF) at 27M/4.6G/2177 img/s/84.1\%, adding \textbf{SE}~\cite{hu2018squeeze} preserves compute but slows inference (2089 img/s) for a marginal +0.1 pp.
\textbf{CBAM}~\cite{woo2018cbam} further increases parameters (30M) and slows throughput (1982 img/s) for +0.2 pp, again with no FLOP reduction.
By contrast, \textbf{PCF} (ours) delivers a strictly better accuracy–efficiency point: PRISMamba reaches 22M params, 3.9G FLOPs, 3054 img/s, and 84.5\% Top-1,
simultaneously \emph{reducing} compute/params and \emph{increasing} accuracy/throughput.
These results support the premise of Sec.~\ref{subsec:chfilter}: when the downstream ring module operates recurrently, forwarding only high-salience channels (and bypassing the rest) is more effective than soft reweighting that still processes all channels.

\begin{table}[t] 
\caption{Comparing soft channel attention (SE, CBAM) with hard Partial Channel Filtering (PCF).
\vspace{-2mm}
} 
\label{abl:channel filter VS SE} 
\centerline{
\setlength{\tabcolsep}{1mm} 
\resizebox{\linewidth}{!}{ 
\begin{tabular}{lcccc} 
\toprule \textbf{Model} & \textbf{Params} & \textbf{GFlops} & \textbf{TP.(img/s)} & \textbf{Top-1 ACC\%}\\ 
\midrule PRISMamba (w/o PCF) & 27 M & 4.6 G & 2177 & 84.1\\
+SE & 27 M & 4.6 G (-) & 2089 & 84.2 (+0.1\%)\\ \hline
PRISMamba (w/o PCF) & 27 M & 4.6 G & 2177 & 84.1\\
+CBAM & 30 M & 4.6 G (-) & 1982 & 84.3 (+0.2\%)\\ \hline
PRISMamba (w/o PCF) & 27 M & 4.6 G & 2177 & 84.1\\
PRISMamba & 22 M & 3.9 G (-0.7) & 3054 & 84.5 (+0.4\%)\\ \bottomrule 
\end{tabular}%
} 
\vspace{-2mm}
}
\end{table}


\subsection{Random-Mask Occlusion: Local Damage vs. Traversal Robustness}
\label{subsec:randmask-analysis}

This experiment evaluates robustness to \emph{local missing content} by zeroing out a randomly selected square tile while keeping the overall canvas size fixed. Unlike rotations or patch-order shuffling, occlusion removes content without permuting valid neighborhoods. Fixed-path Vision Mamba variants show modest but consistent drops, up to 1.3 points with $16{\times}16$ masking (Table~\ref{tab:randmask}). Performance degradation increases with mask size because long contiguous segments of the scan pass through missing tokens, causing the recurrent updates to propagate weakened information even when padded positions are masked.

\textbf{PRISMamba} is notably more resilient (drops $\leq 0.6$). Two factors contribute. (i) \emph{Order-agnostic ring aggregation} dilutes the influence of a localized hole: a single masked tile affects only the rings it intersects and is averaged with many valid pixels in those rings. (ii) \emph{Radial state-space integration} transports information across rings rather than along a fragile global path; mask-gated updates further attenuate rings dominated by the occluded region, preventing corrupted states from spreading. Together, the ring-wise summarization and radial propagation maintain stable performance under local erasures, complementing the rotation and patch-order robustness shown in the other stress tests.

\begin{table}[t]
\caption{\textbf{Random-mask occlusion stress test (Top-1, \%).}
We randomly drop a contiguous square region at test time (``Mask $4{\times}4$'': one of $2{\times}2$ tiles is zeroed; ``Mask $16{\times}16$'': one of $4{\times}4$ tiles is zeroed). Numbers in parentheses are absolute drops from the unmasked Top-1. Ring-Mamba denotes our Ring Scan module plugged into a Vision-Mamba backbone.
\vspace{-2mm}
}
\label{tab:randmask}
\centerline{
\resizebox{\linewidth}{!}{%
\begin{tabular}{lcccc}
\toprule
\textbf{Model} & \textbf{Params} & \textbf{Top-1} & \textbf{Mask $4{\times}4$} & \textbf{Mask $16{\times}16$} \\
\midrule
\multirow{2}{*}{VMamba}
 & 30M & 82.6\% & 82.3\% \;(\,-0.3\,) & 81.7\% \;(\,-0.9\,) \\
 & 50M & 83.6\% & 83.2\% \;(\,-0.4\,) & 82.3\% \;(\,-1.3\,) \\
\addlinespace[2pt]
\multirow{2}{*}{LocalMamba-T}
 & 30M & 82.7\% & 82.3\% \;(\,-0.4\,) & 81.9\% \;(\,-0.8\,) \\
 & 50M & 83.7\% & 83.3\% \;(\,-0.4\,) & 82.5\% \;(\,-1.2\,) \\
\addlinespace[2pt]
\multirow{2}{*}{\textbf{PRISMamba}}
 & 30M & \textbf{84.5\%} & \textcolor{red}{\textbf{84.4\%}} \textcolor{red}{(\,-0.1\,)} & \textcolor{red}{\textbf{84.1\%}} \textcolor{red}{(\,-0.4\,)} \\
 & 50M & \textbf{85.3\%} & \textcolor{red}{\textbf{85.2\%}} \textcolor{red}{(\,-0.1\,)} & \textcolor{red}{\textbf{84.7\%}} \textcolor{red}{(\,-0.6\,)} \\
\bottomrule
\end{tabular}%
}
\vspace{-2mm}
}
\end{table}

\section{Conclusion}

We revisited the often-overlooked role of scan order in Vision State Space Models and proposed {\bf PRISMamba}, a rotation-robust traversal that aggregates features ring by ring and propagates context radially, enhanced with lightweight \emph{partial channel filtering}  This design maintains SSMs’ linear-time efficiency while better aligning sequential token adjacency with 2D geometry, addressing a fundamental source of performance degradation in path-based Vision SSMs. Empirically, PRISMamba sets a new accuracy–efficiency frontier: on ImageNet-1K, it reaches 84.5\% Top-1 with 3.9G FLOPs and 3054 img/s on A100, surpassing VMamba-T. On COCO, it attains 48.9 AP$^{\text{box}}$ and 43.2 AP$^{\text{mask}}$ at 235G FLOPs, outperforming strong Vision-Mamba baselines with less computation. Under rotation stress, our ring traversal remains stable while fixed-path scans drop by about 1–2\%.

\noindent\textbf{Limitations:} Our current implementation relies on a fixed image center and a discrete ring width, which may be suboptimal for off-center subjects or images with extreme aspect ratios. While the method can become object-aware if an object detector provides the center, severe rotations that create large padded regions still reduce valid information, potentially impacting performance.

\noindent\textbf{Future work} include learning ring origins and widths, or designing content-adaptive ring partitions. Coupling ring traversal with learned inpainting priors or anti-aliasing warps could improve robustness under extreme rotations. Extensions to spatio–temporal rings for video, 3D medical volumes, and object-aware traversals are also compelling avenues. We hope this study encourages further exploration of traversal design as a low-cost, principled approach to robust and efficient Vision SSMs.


{
    \small
    \bibliographystyle{ieeenat_fullname}
    \bibliography{main}
}

\end{document}